\documentclass[acmsmall, prodmode,acmtecs]{acmart}

\usepackage[export]{adjustbox}
\usepackage{graphicx}
\usepackage{subcaption}
\usepackage{multirow}
\AtBeginDocument{%
  \providecommand\BibTeX{{%
    \normalfont B\kern-0.5em{\scshape i\kern-0.25em b}\kern-0.8em\TeX}}}

\setcopyright{acmcopyright}
\copyrightyear{2020}
\acmYear{2020}
\acmDOI{10.1145/1122445.1122456}

\acmJournal{JACM}
\acmVolume{37}
\acmNumber{4}
\acmArticle{111}
\acmMonth{8}

\begin{document}

\title{CrowdMOT: Crowdsourcing Strategies for Tracking Multiple Objects in Videos }

\author[1]{Samreen Anjum}
\email{samreen@utexas.edu}
\affiliation{\institution{School of Information, University of Texas at Austin}}
\author[2]{Chi Lin}
\email{benny@houzz.com}
\affiliation{\institution{Houzz, Inc.}}
\author[3]{Danna Gurari}
\email{danna.gurari@ischool.utexas.edu}
\affiliation{\institution{School of Information, University of Texas at Austin}}

\renewcommand{\shortauthors}{Anjum, et al.}

\begin{abstract}
Crowdsourcing is a valuable approach for tracking objects in videos in a more scalable manner than possible with domain experts.  However, existing frameworks do not produce high quality results with non-expert crowdworkers, especially for scenarios where objects split.  To address this shortcoming, we introduce a crowdsourcing platform called CrowdMOT, and investigate two micro-task design decisions: (1) whether to decompose the task so that each worker is in charge of annotating all objects in a sub-segment of the video versus annotating a single object across the entire video, and (2) whether to show annotations from previous workers to the next individuals working on the task.  We conduct experiments on a diversity of videos which show both familiar objects (aka - people) and unfamiliar objects (aka - cells).  Our results highlight strategies for efficiently collecting higher quality annotations than observed when using strategies employed by today's state-of-art crowdsourcing system.  
\end{abstract}

\begin{CCSXML}
<ccs2012>
<concept>
<concept_id>10002951.10003260.10003282.10003296</concept_id>
<concept_desc>Information systems~Crowdsourcing</concept_desc>
<concept_significance>500</concept_significance>
</concept>
<concept>
<concept_id>10010147.10010178.10010224</concept_id>
<concept_desc>Computing methodologies~Computer vision</concept_desc>
<concept_significance>500</concept_significance>
</concept>
<concept_id>10010147.10010257</concept_id>
<concept_desc>Computing methodologies~Machine learning</concept_desc>
<concept_significance>300</concept_significance>
</concept>
</ccs2012>
\end{CCSXML}

\ccsdesc[500]{Information systems~Crowdsourcing}
\ccsdesc[500]{Computing methodologies~Computer vision}

\keywords{Crowdsourcing,Computer Vision,Video Annotation}

% \caption{Seattle Mariners at Spring Training, 2010.}

\maketitle

\section{Introduction}
Videos provide a unique setting for studying objects in a temporal manner, which cannot be achieved with 2D images.  They reveal each object's actions and interactions, which is valuable for applications including self-driving vehicles, security surveillance, shopping behavior analysis, and activity recognition.  Videos also are important for biomedical researchers who study cell lineage to learn about processes such as viral infections, tissue damage, cancer progression, and wound healing.  

\begin{figure*}[t]
    \centering
    \includegraphics[width = 1\textwidth]{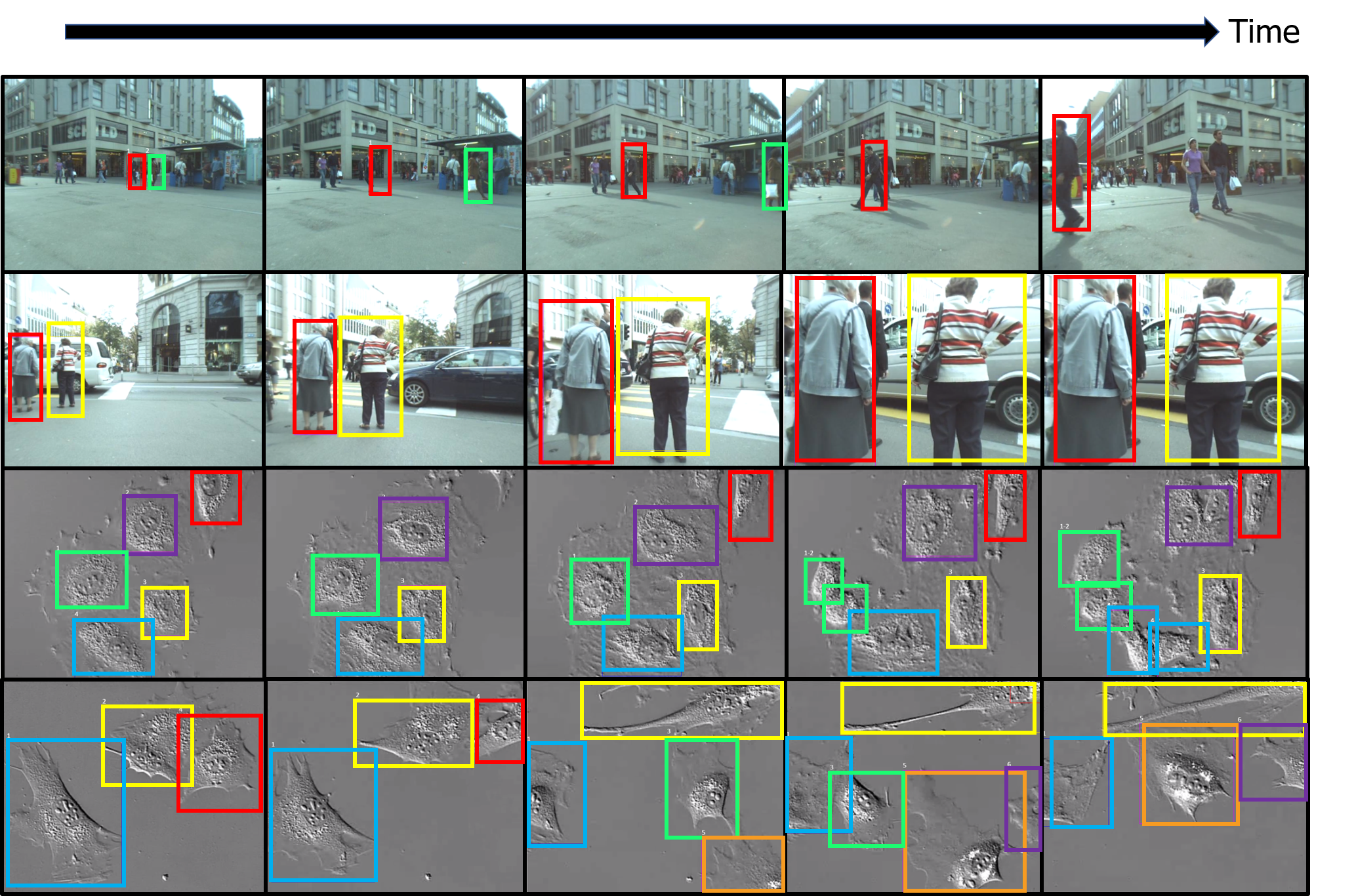}
    \caption{Examples of multiple object tracking (MOT) results collected with our CrowdMOT crowdsourcing platform.  As shown, CrowdMOT supports content ranging from familiar objects such as people (top two rows) to unfamiliar objects such as cells (bottom two rows).  It handles difficult cases including when an object leaves the field of view (first row, green box), leaves the field of view and reappears (third row, red box), appears in middle of the video (fourth row, orange box), changes size dramatically over time (second row), or splits as the cell undergoes mitosis (third row, green box). (best viewed in color)}
    \label{dataset}
\end{figure*}

Many data annotation companies have emerged to meet the demand for high quality, labelled video datasets~\cite{claysciences, playment,figure-eight,scale,Cloudfactory,alegion}.  Some companies employ in-house, trained labellers, while other companies employ crowdsourcing strategies.  Despite their progress that is paving the way for new applications in society, their methodologies remain proprietary.  In other words, potentially available knowledge of how to successfully crowdsource video annotations is not in the public domain.  Consequently, it is not clear whether such companies' successes derive from novel crowdsourcing interfaces versus novel worker training protocols versus other mechanisms.

Towards filling this gap, we focus on identifying successful crowdsourcing strategies for video annotation in order to establish a scalable approach for tracking objects. A key component in analyzing videos is examining how each object behaves over time.  Commonly, it is achieved by localizing each object in the video (detection) and then following all objects as they move (tracking).  \textcolor{black}{This task is commonly referred to as \emph{multiple object tracking (MOT)} \cite{milan2016mot16}}.  One less-studied aspect of MOT is the fact that an object can split into multiple objects.  This can arise, for example, for exploding objects such as ballistics, balloons, or meteors, and for self-reproducing organisms such as cells in the human body (exemplified in Figure~\ref{dataset}).  We refer to the task of tracking all fragments coming from the original object as \emph{lineage tracking}.

While crowdsourcing exists as a powerful option for leveraging human workers to annotate a large number of videos~\cite{vondrick_efficiently_2013,yuen2009labelme},  existing crowdsourcing research about MOT has two key limitations.  First, our analysis shows that today's state-of-art crowdsourcing system and its employed strategies~\cite{vondrick_efficiently_2013} do not consistently produce high quality results with non-expert crowdworkers (Sections~\ref{sec:vatic_evaluation} and \ref{sec:singseg_vs_singobj}). As noted in prior work~\cite{vondrick_efficiently_2013}, the success likely stems from employing expert workers identified through task qualification tests, a step which reduces the worker pool and so limits the extent to which such approaches can scale up. 
Second, prior work has only evaluated MOT solutions for specific video domains; e.g., only videos showing \emph{familiar content} like people~\cite{vondrick_efficiently_2013} or only videos showing \emph{unfamiliar content} like biological cells~\cite{sameki_crowdtrack:_2016}.  This begs a question of how well MOT strategies will generalize across such distinct video domains, which can manifest unique annotation challenges such as the need for lineage tracking.

 To address these concerns, we focus on (1) proposing strategies for decomposing the complex MOT into microtasks that can be completed by non-expert crowdworkers, and (2) supporting MOT annotation for both familiar (people) and unfamiliar (cell) content, thereby bridging two domains related to MOT.

We analyze two crowdsourcing strategies for collecting MOT annotations from a pool of non-expert crowdworkers for both familiar and unfamiliar video content. First, we compare two choices for decomposing the task of tracking multiple objects in videos, i.e., track all objects in a segment of the video (time-based approach that we call SingSeg) or track one object across the entire video (object-based approach that we call SingObj). Second, we examine if \textcolor{black}{creating \textit{iterative tasks}, where crowdworkers see} the results from a previous worker on the same video, improves annotation performance. 
 
 To evaluate these strategies, we introduce a new video annotation platform for MOT, which we call \textit{CrowdMOT}. CrowdMOT is designed to support lineage tracking as well as to engage non-expert workers for video annotation. Using CrowdMOT, we conduct experiments to quantify the efficacy of the two aforementioned design strategies when crowdsourcing annotations on videos with multiple objects. Our analysis with respect to several evaluation metrics on diverse videos, showing people and cells, highlights strategies for collecting much higher quality annotations from non-expert crowdworkers than is observed from strategies employed by today's state-of-the art system, VATIC~\cite{vondrick_efficiently_2013}. 
 
 To summarize, our main contribution is a detailed analysis of two crowdsourcing strategy decisions: (a) which microtask design and (b) whether to use an iterative task design.  Studies demonstrate the efficacy of these strategies on a \textcolor{black}{variety} of videos showing familiar (people) and unfamiliar (cell) content.  Our findings reveal which strategies result in higher quality results when collecting MOT annotations from non-expert crowdworkers.  We will publicly-share the crowdsourcing system, CrowdMOT, that incorporates these strategies.
\section{Related Work}
 \vspace{0.5em}
 \subsection{Crowdsourcing Annotations for Images \textit{vs.} Videos}  
 Since the onset of crowdsourcing, much research has centered on annotating visual content.  Early on, crowdsourcing approaches were proposed for simple tasks such as tagging objects in images~\cite{vonahn2004Labelingimagescomputer}, localizing objects in images with bounding rectangles~\cite{vonahn2006Peekaboomgamelocating}, and describing salient information in images~\cite{vonahn2006Improvingaccessibilityweb,salisbury2017toward,kohler2017supporting}. More recently, a key focus has been on developing crowdsourcing frameworks to address more complex visual analysis tasks such as counting the number of objects in an image~\cite{sarma2015surpassing}, creating stories to link collections of distinct images~\cite{Mandal2017collective}, critiquing visual design~\cite{luther2014crowdcrit}, investigating the dissonance between human and machine understanding in visual tasks~\cite{zhang2019dissonance}, and tracking all objects in a video~\cite{vondrick_efficiently_2013}.  Our work contributes to the more recent effort of developing strategies to decompose complex visual analysis tasks into simpler ones that can be completed by non-expert crowdworkers.  The complexity of video annotation arises in part from the large amount of data, since even small videos consist of several thousand images that must be annotated; e.g., a typical one-minute video clip contains 1,740 images.  Our work offers new insights into how to collect high quality video annotations from an anonymous, non-expert crowd.
 
\subsection{Crowdsourcing Video Annotations} 
Within the scope of crowdsourcing video annotations, there are a broad range of valuable tasks. Some crowdsourcing platforms promote learning by improving content of educational videos~\cite{cross2014vidwiki} and editing captions in videos to learn foreign languages~\cite{culbertson2017have}. Other crowdsourcing systems employ crowdworkers to flag frames where events of interest begin and/or end.  Examples include activity recognition \cite{nguyen2013tagging}, event detection \cite{steiner2011crowdsourcing}, behavior detection \cite{park2012crowdsourcing}, and more~\cite{abu2016youtube,heilbron_activitynet:_2015}. Our work most closely relates to the body of work that requires crowdworkers to not only identify frames of interest in a video, but also to localize all objects in every video frame~\cite{vondrick_efficiently_2013,yuen2009labelme,sameki_crowdtrack:_2016}.  The most popular ones that complete this MOT task include VATIC \cite{vondrick_efficiently_2013} and LabelMe Video \cite{yuen2009labelme}.  In general, these tools exploit temporal redundancy between frames in a video to reduce the human effort involved by asking users to only annotate key frames, and have the tool interpolate annotations for the intermediate frames~\cite{vondrick_efficiently_2013,yuen2009labelme}.  Our work differs in part because we propose a different strategy for decomposing the task into microtasks.  Our experiments on \textcolor{black}{videos showing both familar and unfamiliar content} demonstrate the advantage of our strategies over strategies employed in today's state-of-the-art crowdsourcing system~\cite{vondrick_efficiently_2013}.

\subsection{Task Decomposition}
One of the key components in effective crowdsourcing is to divide large complex tasks into smaller atomic tasks called microtasks. These atomic or unit tasks are typically designed in such a way that they pose minimal cognitive and time load. Decomposing tasks into microtasks can lead to faster results (through greater parallelism) \cite{bernstein2010soylent,little2009turkit,kittur2011crowdforge,latoza2013crowd,tong2018slade} with higher quality output \cite{cheng2015break,teevan2016productivity}. Effective microtasks have been applied, for example, to create taxonomies \cite{chilton2013cascade}, generate action plans \cite{kaur2018creating}, construct crowdsourcing workflows \cite{kulkarni2012collaboratively} and write papers \cite{bernstein2010soylent}. \textcolor{black}{Within visual content annotation, several strategies combining human and computer intelligence have been designed to localize objects in difficult images \cite{russakovsky2015best}, segment images \cite{song2017tool} and reconstruct 3D scenes \cite{song2019popup}. Additionally, workflows have been proposed to efficiently geolocate images by allowing experts to work with crowdworkers  \cite{venkatagiri2019groundtruth}. }  Unlike prior work, we focus on effective task decomposition techniques for the MOT problem.  Our work describes the unique challenges of this domain (i.e., spatio-temporal problem to follow objects spatially and temporally across large number of frames) and provides a promising microtask solution. 

\subsection{Iterative Crowdsourcing Tasks}
Crowdsourcing approaches can be broadly divided into two types: \textit{parallel}, in which workers solve a problem alone, and \textit{iterative}, in which workers serially build on the results of other workers \cite{little2010exploring}. Examples of iterative tasks include interdependent tasks~\cite{kim2017mechanical} and engaging workers in multi-turn discussions  \cite{chen2019cicero}, which can lead to improved outcomes such as increased worker retention \cite{gadiraju2017improving}.  Prior work has also demonstrated workers perform better on their own work after reviewing others' work \cite{kobayashi2018empirical, zhu2014reviewing}.  The iterative approach has been shown to produce better results for the tasks of image description, writing, and brainstorming \cite{little2010exploring, little2009turkit,zhang2012human}.  More recently, an iterative process has been leveraged to crowdsource complex tasks such as masking private content in images \cite{kaur2017crowdmask}.  Our work complements prior work by demonstrating the advantage of exposing workers to previous workers' high quality annotations on the same video in order to obtain higher quality results for the MOT task.
 
\subsection{Tracking Cells in Videos}  
As evidence of the importance of the cell tracking problem, many publicly-available biological tools are designed to support this type of video annotation: CellTracker \cite{piccinini_celltracker_2016}, TACTICS \cite{shimoni_tactics_2013}, BioImageXD \cite{kankaanpaa_bioimagexd:_2012}, eDetect \cite{han_edetect:_2019}, LEVER \cite{winter_lever:_2016}, tTt \cite{hilsenbeck_software_2016}, NucliTrack \cite{cooper_nuclitrack:_2017}, TrackMate \cite{tinevez_trackmate:_2017}, and Icy \cite{de_chaumont_icy:_2012}. However, only one tool~\cite{sameki_crowdtrack:_2016} is designed for use in a crowdsourcing environment, and it was evaluated for tracking cells based on more costly object segmentations (rather than less costly, more efficient bounding boxes).  Our work aims to bridge this gap by not only seeking strategies that work for cell annotation but also generalize more broadly to support videos of familiar everyday content. CrowdMOT is designed to support cell tracking, because it features lineage tracking by recognizing when a cell undergoes mitosis and so splits into children cells (exemplified in Figure~\ref{dataset}, row 3).

\section{Pilot Study: Evaluation of State-of-Art Crowdsourcing System}
\label{sec:vatic_evaluation}
\textcolor{black}{Our work was inspired by our observation that we obtained poor quality results when we used today's state-of-art crowdsourcing system, which is called VATIC~\cite{vondrick_efficiently_2013}, to employ non-expert crowdworkers to track biological cells.  Based on this initial observation, we conducted follow-up experiments to assess the reasons for the poor quality results.  We chose to conduct these and subsequent experiments on both familiar everyday content and unfamiliar biological content showing cells in order to ensure our findings represent more generalized findings.}

\subsection{Experimental Design} 

\paragraph{Dataset.} 
We conducted experiments with 35 videos containing 49,501 frames showing both familiar content (people) and unfamiliar content (cells). Of these, 15 videos (11,720 frames) show people \footnote{These videos came from the MOT dataset: https://motchallenge.net/} and the remaining 20 videos (37,781 frames) show cells \footnote{These videos came from the CTMC dataset \cite{Anjum_2020_CVPR_Workshops}.
We collected ground truth data for all videos from two in-house experts \footnote{Our experts were two graduate students who had successfully completed a course about crowdsourcing visual content.} who we trained to perform video annotation.  }. 

\paragraph{VATIC Configuration.}
\textcolor{black}{
We collected annotations with the default parameters, where each video was split into smaller segments of 320 frames with 20 overlapping frames.  This resulted in a total of 181 segments. A new crowdsourcing job was created for each segment and assigned to a crowdworker.  VATIC then merged the tracking results from consecutive segments using the Hungarian algorithm \cite{munkres1957algorithms}. }

\textcolor{black}{The VATIC instructions indicate to mark all objects of interest and to mark one object at a time in order to avoid confusion. That is, workers were asked to complete annotating one object across the entire video, and then rewind the video to begin annotating the next object. For each object, workers were asked to draw the rectangle tightly such that it completely encloses the object.} 

To assist workers with tracking multiple objects, the interface enabled them to freely navigate between frames to view their annotations at any given time. Each object is marked with a unique color along with a unique label on the top right corner of the bounding box to visually aid the worker with tracking that object. 

Our implementation had only one difference from that discussed in prior work~\cite{vondrick_efficiently_2013}.  We did not filter out workers using a ``gold standard challenge".  The original implementation, in contrast, prevented workers from completing the jobs unless they passed an initial annotation test. 

\paragraph{Crowdsourcing Environment and Parameters.}
\textcolor{black}{
As done for the original evaluation of VATIC~\cite{vondrick_efficiently_2013}, we employed crowdworkers from Amazon Mechanical Turk (AMT).  We restricted the jobs to workers who had completed at least 500 Human Intelligence Tasks (HITs) and had at least a 95\% approval rating. We paid \$0.50 per HIT and assigned 30 minutes to complete each HIT. \footnote{Of note, we conducted this experiment before June 2019, and since then, the AMT API used by the VATIC system has been deprecated, rendering VATIC incompatible for crowdsourcing with AMT.}  }

\paragraph{Evaluation Metrics.}
\textcolor{black}{
We compared the results obtained using the VATIC system with the ground truth data and evaluated the tracking performance with commonly used metrics for object tracking tasks \cite{wu2013online}. Specifically, we employed the three following metrics: 
\begin{enumerate}
    \item Area Under Curve (AUC) measures the accuracy of the size of the bounding boxes
    \item Track Accuracy (TrAcc) measures the number of frames in which the object was correctly annotated 
    \item Precision measures the accuracy of the central location of the bounding boxes
\end{enumerate}
For all these metrics, the resulting values range from 0 to 1 with higher values indicating better performance. Further description on how these metrics are computed is provided in Section \ref{section: experiments}.
}
\subsection{Results} 
Overall, we observed poor quality results, as indicated by low scores for all three metrics: AUC is 0.06, TrAcc is 0.42, and Precision is 0.03.  This poor performance was surprising to us given VATIC's popularity.  For example, it has been reportedly used to generate several benchmark datasets as recently as 2018 \cite{muller2018trackingnet,russakovsky2015imagenet}. We reached out to one of the authors~\cite{muller2018trackingnet}, who clarified that even for annotation of videos with single objects, they added a significant amount of quality assurance and microtask design modifications to the system in order to collect high quality annotations. For example, they reported that they hired master workers and underwent several rounds of verification and correction by both crowdworkers and experts.  

In what follows, we identify reasons for the poor performance of VATIC.  We also introduce a new system to try to address VATIC's shortcomings while building on its successes.  We will demonstrate in Section \ref{section: experiments} that modification of crowdsourcing strategies employed in VATIC leads to improved tracking performance.  We refer the reader to the Appendix for a direct comparison between our new system and VATIC.
\section{CrowdMOT}
\label{Approach}
\begin{figure}[t!]
    \centering
    \frame{\includegraphics[width=1\textwidth]{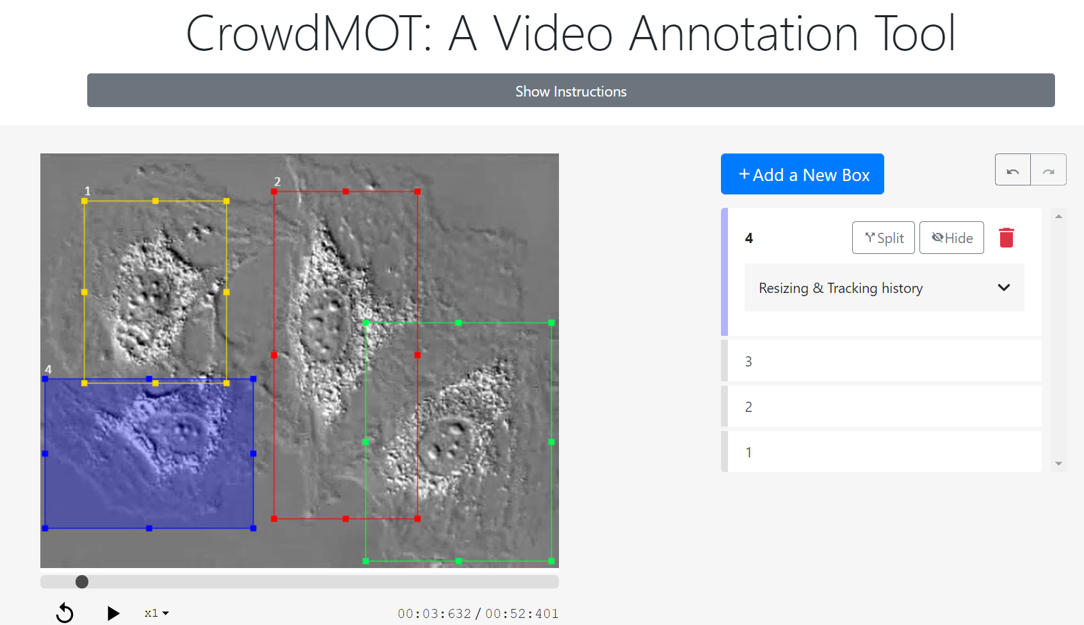}}
    \caption{User interface of our crowdsourcing system. The interface dynamically loads videos from a URL. Users can then draw and resize bounding boxes to detect and track multiple objects in a video. Users can also  perform lineage tracking for objects that \textit{split}. Users can play or adjust the speed of the video, and replay the video to view or adjust the interpolated annotations.}
    \label{crowdmot-eg}
\end{figure}

\textcolor{black}{We now introduce CrowdMOT, a web-based crowdsourcing platform for video annotation that supports lineage tracking. Our objective was to improve upon the basic user interface and crowdsourcing strategies adopted for VATIC~\cite{vondrick_efficiently_2013}, preserving the targeted support for videos showing familiar content while extending it to also support videos showing migrating cells.  A screen shot of CrowdMOT's user interface is shown in Figure~\ref{crowdmot-eg}.  In what follows, we describe the evolution of our CrowdMOT system in order to highlight the motivation behind our design choices for various features included in the system.}

\subsection{Implementation Details} 
\textcolor{black}{We began by migrating the outdated code-base for VATIC~\cite{vondrick_efficiently_2013} into a more modern, user-friendly system, which we call CrowdMOT.  We developed the system using React, Konva, and javascript. React is a simple and versatile javascript framework. Amongst it many advantages, it loads webpages quickly and supports code reusability for simplified development and extensions.} Konva is a library that supports easy integration of shapes. Finally, javascript integrates well with modern browsers. 

\subsection{Process Flow} 
With CrowdMOT, users are required to follow two key steps similar to VATIC.  First, a user draws a bounding box around an object to begin annotation by \textit{clicking and dragging} a new bounding box around it, with eight adjustable points that can be moved to tighten the box's fit around the object. While VATIC's bounding boxes have adjustable edges requiring users to move two edges for resizing, we made a minor change to reduce human effort by providing eight adjustable points that allows users to resize the box using one point. Second, the user moves the box to track the object.  To do so, the user plays the video and, at any time, pauses it to relocate and refit the bounding box to a new location for the object.  Following prior work \cite{vondrick_efficiently_2013}, the user adjusts each bounding box only in a subset of key frames, and the box is propagated between the intermediate frames using linear interpolation.  

\begin{figure*}[t!]
    \centering
    \includegraphics[width = 1\textwidth]{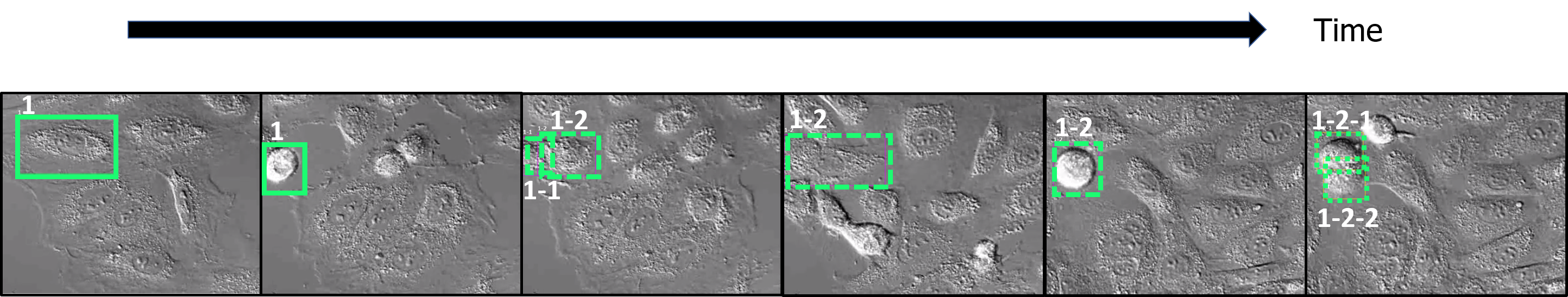}
    \caption{\textbf{Lineage tracking. }Example of CrowdMOT handling a case when a cell undergoes two rounds of consecutive mitosis (cell division). The bounding boxes are shown as dotted lines to depict the splitting event. Each image represents frames extracted from the video at different time stamps. In the first image, the user marks a cell labelled 1. This cell undergoes mitosis as shown in the third image, and splits into two children cells labelled 1-1 and 1-2. In the fourth image, child cell 1-1 leaves the video frame and only child cell 1-2 remains. This child cell then undergoes mitosis again in the last image, which results in the creation of two new children cells, 1-2-1 and 1-2-2. }
    \label{splitExample}
\end{figure*}

%[New]
Extending VATIC, to support lineage tracking, users of CrowdMOT also can mark a split event at any frame.  When a user flags a frame where this event occurs, the existing box splits into two new boxes, which can be adjusted by the user to tightly fit around the new objects.  An example is illustrated in Figure \ref{splitExample}.  The system also records for each split object its childrens' ids and parent's id, if available, to support lineage tracking.  

\subsection{User Interface} 
We introduced the following user interface choices, drawing inspiration both from prior work~\cite{vondrick_efficiently_2013} and our pilot studies. 

\begin{enumerate}
   \item\textit{Instructions and How-to Video.} Given the significance of the quality of instructions in crowdsourcing tasks \cite{wu2017confusing}, we performed iterative refinements of the instructions through pilot studies. We ultimately provided a procedural step by step format, with accompanied video clips demonstrating how to annotate a single object as well as how to use specific features such as flagging when an object undergoes splitting.

\item\textit{Labeling.} Inspired from prior work~\cite{vondrick_efficiently_2013}, we restricted the user interface to keep it simple by neither allowing users to enter free text to label the objects nor allowing users to provide any free style shapes to draw around objects. This is exemplified in Figure \ref{crowdmot-eg}. We provided unique identifiers for each object to help users keep track of the annotations. In addition, to retain visual connection and keep track of each parent object's progeny, each new child object is labelled with a new unique identifier that includes the parent's unique identifier. 

\item\textit{Playback.} Similar to prior work~\cite{vondrick_efficiently_2013}, users can play and replay the video in order to preview the interpolated annotations and make any edits to the user-defined or interpolated frames. 

\item\textit{Speed Control.} Since  individual  learning  and  video content absorption level varies person by person, following prior work~\cite{vondrick_efficiently_2013}, we included this feature that allows users to change the playback speed of the video by slowing it down or speeding it up as deemed appropriate for efficient annotation \cite{caruso2016slow}.

\item\textit{Preview.} Inspired by our pilot studies, we added a new feature in CrowdMOT to enforce workers to review their final work. When a worker is ready to submit results, the worker must review the entire video with the interpolated annotation to verify its quality before submission.

\item\textit{Feedback.} We  also  introduced  a  feedback  module  for  soliciting feedback or suggestions from the workers about the task.

\end{enumerate}

After pilot tests with this infrastructure in place, we observed very poor quality results. Upon investigation, we initially attributed this to the following two issues which led us to make further system improvements:

\begin{enumerate}
    \item\textit{Key Frames.} 
In order to take advantage of temporal redundancy and reduce user effort, prior work suggests requiring users to move the box that is tracking an object only at fixed key frames~\cite{vondrick_efficiently_2013}.  \textcolor{black}{While this technique has been shown to be faster for annotating familiar objects such as people and vehicles, it} can be a poor fit for when objects split, such as for biological videos showing cells that undergo mitosis.  Using fixed key frames may lead the user to miss the frame where the split occurs, resulting in incorrect interpolation and possibly mistaken identity.  Consequently, we instead have users pick the frames when they wish to move the box bounding the object.

\item\textit{Quality Control.} Due to many users submitting work without any annotations, we only activate the submit button after the user creates a bounding box and moves it at least once in the video. 
\end{enumerate}

\textcolor{black}{With these enhancements, we conducted another round of pilot tests and only observed incremental improvements. } From visual inspection of the results, we hypothesized that the remaining problems stemmed from the microtask design rather than the task interface. Hence, we propose two alternate crowdsourcing strategies, in the form of task decomposition and \textcolor{black}{iterative tasks}, which we elaborate on in the next section. 

\begin{figure}[b!]
    \centering
    \includegraphics[width = 1\textwidth]{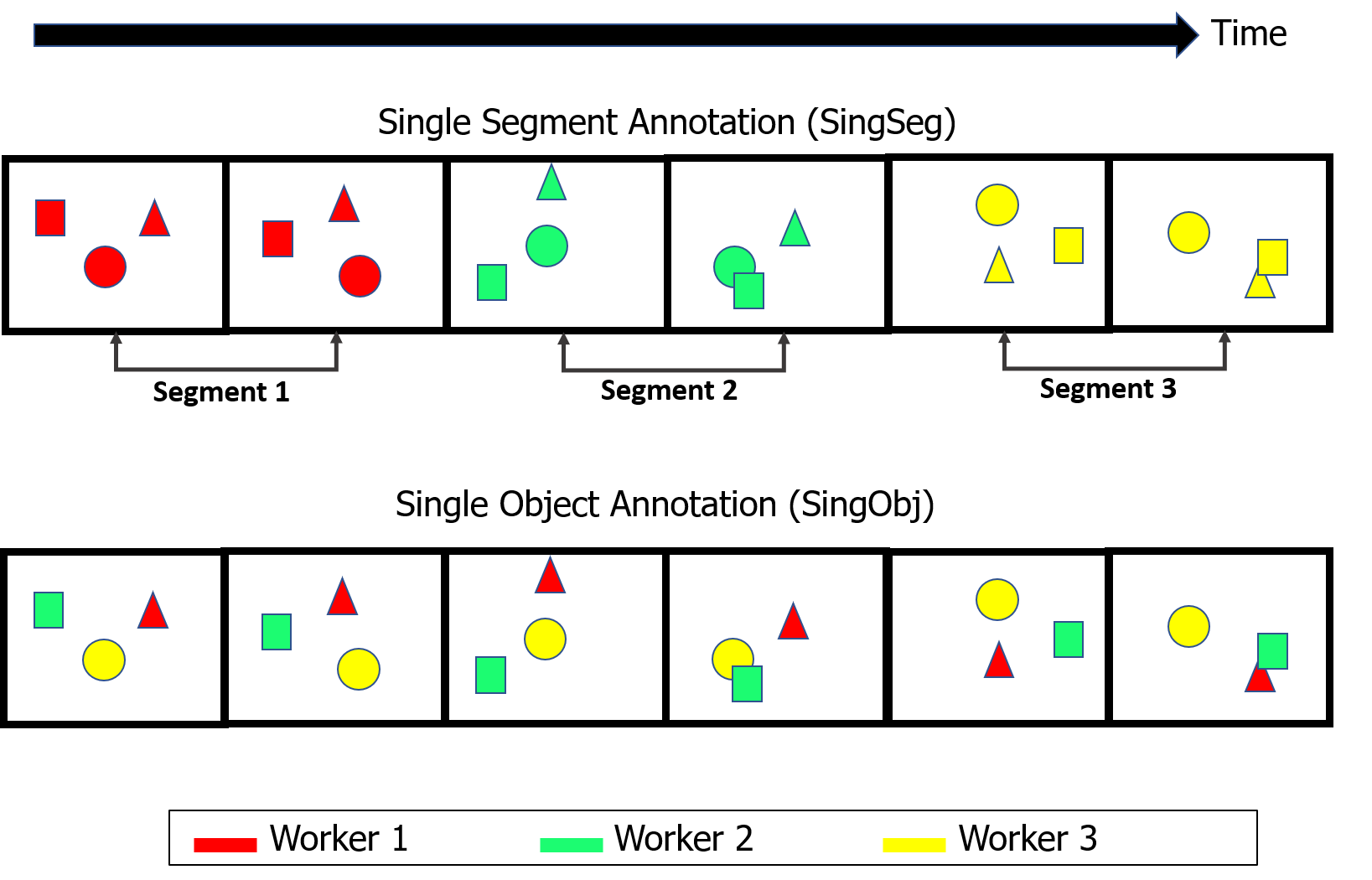}
    \caption{Illustration of two microtask designs for crowdsourcing the MOT task: \emph{Single Segment} (top row) and \emph{Single Object} (bottom row). Each black box represents a frame in a video, each shape represents an object, and each color denotes work to be done by a worker.  A worker is either assigned a full segment of a video for \emph{Single Segment} or a single object for its lifetime in the video for \emph{Single Object}.}
    \label{fig:task}
\end{figure}

\section{Crowdsourcing Strategies for Multiple Object Tracking}
We now describe key choices that we include in CrowdMOT to support collecting higher quality MOT annotations from lay crowdworkers. 

\subsection{Microtask Design Options} 
We introduced two options for how to decompose the task of tracking multiple objects in videos for crowdsourcing, which are illustrated in Figure~\ref{fig:task}.

\textit{Single Segment (SingSeg)}: Prior work~\cite{vondrick_efficiently_2013} recommends a microtask design of splitting the video into shorter segments, and having a single worker annotate all objects in the entire segment one by one. In this paper, we refer to this strategy as \emph{Single Segment (SingSeg)} annotation (exemplified in Figure \ref{fig:task}). However, our pilot studies as well as our first experiment in this paper reveal that we obtain low quality results when following this approach. 

\textit{Single Object (SingObj)}:   Given the limitations of the above approach, we introduced a different microtask design option which limits the number of bounding boxes an individual can draw, and so objects a user can annotate. Specifically, the user is asked to mark only one object. The button that allows users to create a new bounding box is disabled after one box is drawn so that users can mark only one object through its lifetime in the entire video. We refer to this strategy as \emph{Single Object (SingObj)} annotation (exemplified in Figure \ref{fig:task}). \textcolor{black}{Our motivation for designing this strategy can be attributed to our observation during our pilot studies with the VATIC system.  We noticed most users would mark a varying number of objects in a video, and seldom mark all objects. This observation warranted us to redesign a microtask that was simpler and more even across workers.  We also wanted to simplify the cognitive load for users by having them mark one object at a time rather than many at the same time.}  

\subsection{Collaboration Through Iteration }
CrowdMOT enables requesters to import and edit previously completed annotations on the video.  This feature is valuable for collaborative annotation in an iterative manner by allowing subsequent workers to see the annotated objects already completed by previous workers.  We hoped this feature would implicitly deter workers from marking the same object that was previously annotated by other workers.  This feature also could be beneficial for verifying or correcting previous annotations.  

The notion of creating iterative tasks and exposing workers to previous annotations on the same video relates to that discussed in prior work~\cite{ipeirotis2010analyzing,jiang2014efficient}, which distinguishes between microtasks that are dependent (on previous crowdsourced work) and microtasks that are independent.  While independent tasks can be assigned in parallel and merged at the end, dependent tasks instead build on prior annotations. While independent tasks are advantageous in terms of scaling, we will show in our experiments that integrating the iterative, dependent microtask design leads to higher quality results. 

\section{Experiments and Analysis}
\label{section: experiments}
We conduct two studies to explore the following research questions: 

\begin{enumerate}
    \item \textcolor{black}{How does CrowdMOT with SingObj approach compare with its SingSeg counterpart?} 
    \item What impact does collaboration via \textcolor{black}{iteration} have on workers' performance?
\end{enumerate}

\subsection{Study 1: Microtask Comparison of SingSeg \textit{vs}. SingObj}
\label{sec:singseg_vs_singobj}
In this study, we compare the performance of CrowdMOT configured for two microtask designs for tracking multiple objects in videos: SingSeg versus SingObj.  Although prior work \cite{vondrick_efficiently_2013} recommended against the SingObj approach, our observations will lead us to conclude differently. 

\subsubsection{CrowdMOT Configurations:} 
\textcolor{black}{We employed the same general experimental design for both crowdsourcing strategies to enable fair comparison. The key distinction between the two set-ups is the first strategy gives a worker ownership of all objects in a short segment of the video while the second gives a worker ownership of a single object across the entire duration of the video.  More details about each set-up are described below.}

For CrowdMOT-SingSeg, \textcolor{black}{we set the system parameters to match those employed for the state-of-art MOT crowdsourcing environment, VATIC (which employs the SingSeg strategy), by splitting each video into segments of 320 frames with 20 overlapping frames. We create a new HIT for each segment and assign each HIT to three workers. For each segment, out of the three annotations, we picked the one with the highest AUC score as input for the final merge of annotations from consecutive segments. We do so to simulate human supervision of selecting the best annotation. The tracking results are finalized by merging the annotations from consecutive segments using the Hungarian algorithm \cite{munkres1957algorithms}.} 

For CrowdMOT-SingObj, \textcolor{black}{we created a new HIT for each object in an iterative manner.  Paralleling the SingSeg strategy, we assigned each HIT to three workers and picked the resulting highest scoring annotation based on AUC. The instructions specified that workers should annotate an object that is not already annotated.  Of note, workers that marked \textit{split} events were expected to track all subsequent children from the original object.  Evaluation was conducted for all the objects and any of their progeny for all videos.}

\subsubsection{Datasets and Ground Truth:}
 \textcolor{black}{To examine the general-purpose benefit of the strategies we are studying, we chose two sets of videos that are typically studied independently in two different communities (computer vision and biomedical).  We describe each set below.}

We selected 10 videos from the VIRAT video dataset \footnote{\url{https://viratdata.org/}}, which show familiar everyday recordings of pedestrians walking on university grounds and streets. The number of people in these videos vary from 2 to 8 and the videos range from 20 to 54 seconds (i.e., 479 to 1,616 frames).  This dataset is commonly used to evaluate video annotation algorithms in the computer vision community, and presents various challenges in terms of resolution, background clutter, and camera jitter.  The videos are relevant for a wide range of applications, such as video surveillance for security, activity recognition, and human behaviour analysis.

We also selected 10 biomedical videos showing migrating cells from the CTMC dataset \cite {Anjum_2020_CVPR_Workshops}, which supports the biomedical community. The total number of cells in these videos vary from 5 to 11 cells, which includes children cells that appear from a parent cell repeatedly undergoing mitosis (cell division).  The videos range from 45 to 72 seconds (i.e., 1,351 to 2,164 frames). Different types of cells are included that have varying shapes, sizes, and density.  These videos represent important cell lines widely studied in biomedical research to learn about ailments such viral infections, tissue damage, cancer detection, and wound healing.

Altogether, these 20 videos contain 23,938 frames that need to be annotated.  Of these, 6,664 frames come from 10 videos showing people, and 17,274 frames from 10 videos showing live, migrating cells. In total, the videos contained 121 objects, of which 49 belonged to the familiar videos and 79 (combination of parent and children objects) belong to the cell videos. 

For all videos, we collected ground truth data from two in house experts \footnote{Our experts were two graduate students who had successfully completed a course about crowdsourcing visual content, and we trained them to complete video annotation.} who were trained to perform video annotation. This was done as the cell dataset lacked ground truth data. The videos were evenly divided between the annotators. In addition, each annotator rated each video in terms of its difficulty level - easy, medium and hard. The distinction was based on both the time taken to track objects and the complexity of videos in terms of number of objects in each frame.

\subsubsection{Crowdsourcing Parameters:}
We employed crowdworkers from Amazon Mechanical Turk who had completed at least 500 Human Intelligence Tasks (HITs) and had at least a 95\% approval rating. During pilot studies, we found that a worker, on average, takes four minutes to complete a task for both SingObj and SingSeg microtask designs. Based on this observation, we compensated \$0.50 per HIT to pay above minimum wage at an \$8.00 per hour rate. We gave each worker 60 minutes to complete each HIT.  To capture the results from a similar makeup of the crowd, all crowdsourcing was completed in the same time frame (May 2020).

We conducted a between-subject experiment, to minimize memory bias and learning effect in workers that may have otherwise resulted by previous exposure to the videos.  That meant that we ensured we had distinct workers for each CrowdMOT configuration.

\subsubsection{Evaluation Metrics:} To evaluate the tracking performance, we employ
the same commonly used metrics for the evaluation of object tracking tasks that we employed in Section~\ref{sec:vatic_evaluation}. These metrics reflect the performance in terms of the size of the annotation boxes as well as the central location of the boxes \cite{wu2013online}, as described below.  

\emph{Tracking Boxes:} This indicates the ratio of successful frames in which the overlap ratio of the bounding box of the tracked object and the ground truth is higher than a threshold.  A \emph{success plot} is then created by varying the threshold from 0 to 1 and plotting the resulting scores.  The following are concise metrics for characterizing the plot: 
\begin{quote}
\begin{itemize}
\item\emph{Area Under Curve} (AUC): A single score indicating the area underneath the curve in the success plot. The values range from 0 to 1, with 1 being better.
\item\textcolor{black}{\emph{Track Accuracy} (TrAcc)}: A single score indicating the percentage of frames in which a nonzero overlap is sustained between the bounding box and ground truth.  It reflects the accuracy of an object's lifetime in a video. The values range from 0 to 1, and higher values indicate better accuracy.
\end{itemize}
\end{quote}

\emph{Tracking Points:} This indicates the ratio of frames in which the center distance between the bounding box of the tracked object and the ground truth is within a given threshold. The plot, also known as the \emph{precision plot}, is then created by varying the value of threshold from 0 to 50. 

\begin{quote}
\begin{itemize}
    
    \item\emph{\textcolor{black}{Precision}}: A single score from fixing the threshold to the conventional distance of 20 pixels. The scores range from 0 to 1, with 1 being better.
\end{itemize}
\end{quote}

These metrics implicitly measure the success of detecting \emph{split events}.  That is because each object’s lifetime is deemed to end when it splits and two new children are born.  A missed splitting event would mean that the lifetime of the object would last much longer than what is observed in the ground truth, which would lead to low scores for all metrics: AUC, \textcolor{black}{TrAcc} and \textcolor{black}{Precision}.

We also measured the \emph{effort} required by crowdworkers, in terms of the time taken and number of key frames annotated. To calculate the time taken by each worker to complete a HIT, we recorded the time from when the HIT page is loaded until the job is submitted.

\subsubsection{Results - Work Quality:} 
\begin{figure*}[t!]
    \centering
    \includegraphics[width=1\textwidth]{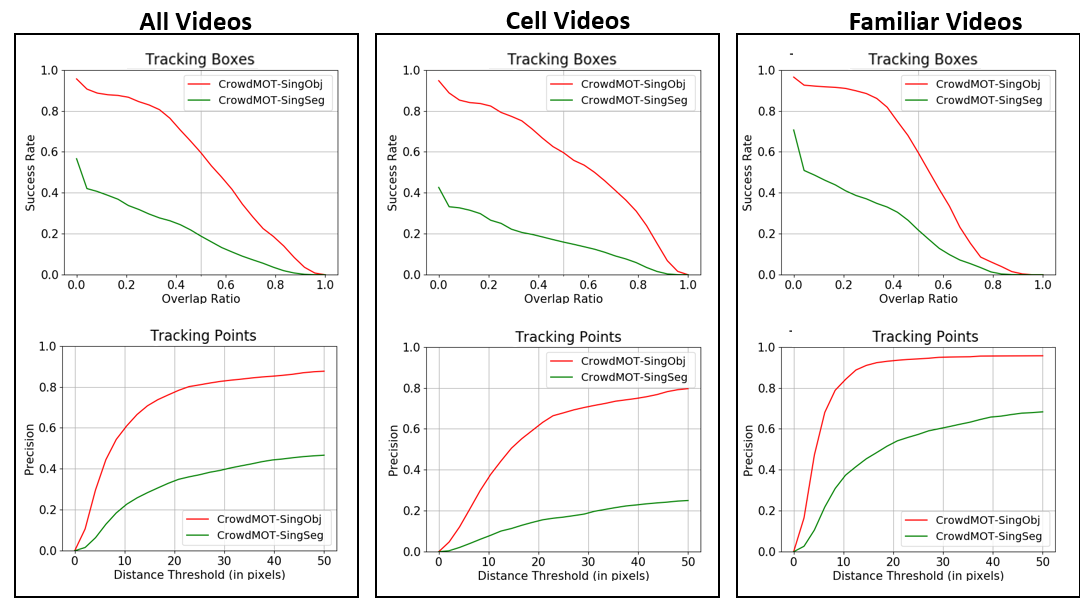}
    %\subfigure{
    %\includegraphics[width=0.32\textwidth]{figures/AllVideos-v3.png}
    %}
    %\subfigure{
    %\includegraphics[width=0.32\textwidth]{figures/CellVideos-v3.png}
    %}
    %\subfigure{
    %\includegraphics[width=0.32\textwidth]{figures/CommonVideos-v3.png}
    %}
    \caption{Success and precision plots comparing SingSeg  and SingObj results. The top row shows results based on evaluating \textit{Tracking Boxes}, and the bottom row is based on \textit{Tracking Points}. CrowdMOT-SingObj annotation approach outperforms CrowdMOT-SingSeg for both familiar pedestrian and unfamiliar cell videos. }
    \label{exp2Fig}
\end{figure*}

The success and precision plots for the crowdsourced results for CrowdMOT-SingSeg and CrowdMOT-SingObj are shown in Figure \ref{exp2Fig} and the average AUC, TrAcc and Precision scores are summarized in Table \ref{study1Tab}. 

We observed poor quality results from \textbf{CrowdMOT-SingSeg} in terms of tracking both boxes and points for all videos.   For instance, as reported in Table \ref{study1Tab}, the overall AUC score for tracking boxes (which summarizes the results in the success plot) is 0.20.  In addition, the {TrAcc} score is 0.57.  This result indicates that, after merging, a considerable portion of frames in which the objects appeared were left unannotated.  Altogether, these findings reveal that the strategy of asking the crowd to annotate segments of videos does not consistently produce high quality annotations.

\begin{figure}[t!]
\begin{minipage}[b]{0.31\textwidth}
\includegraphics[width=1\textwidth]{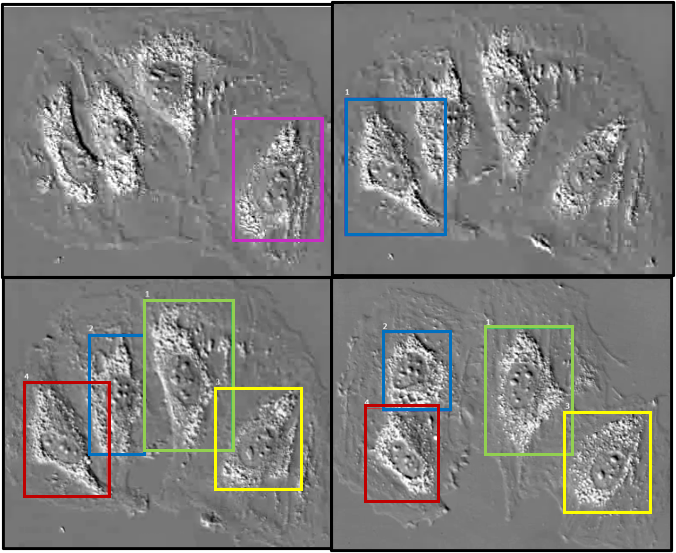}\\
\subcaption{}
\label{study1eg1}
\end{minipage}%
\hspace{0.5pt}
\begin{minipage}[b]{0.35\textwidth}
\includegraphics[width=1\textwidth]{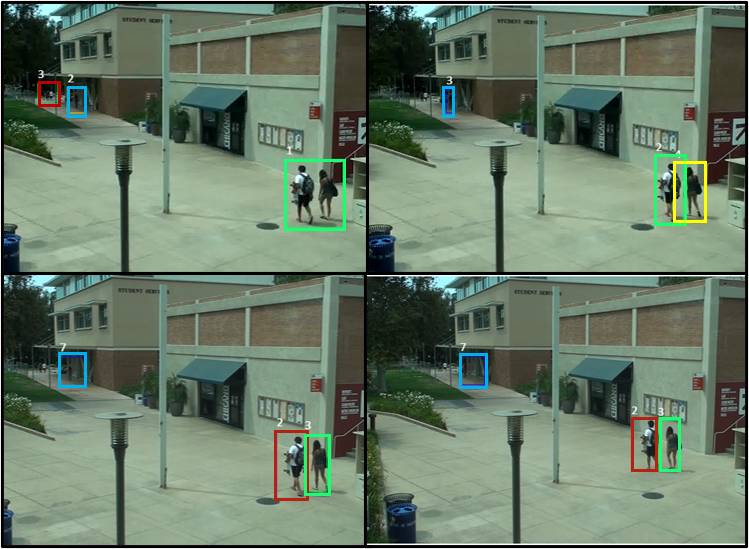}\\
\subcaption{}
\label{study1eg2}
\end{minipage}%
\hspace{0.5pt}
\begin{minipage}[b]{0.32\textwidth}
\includegraphics[width=1\textwidth]{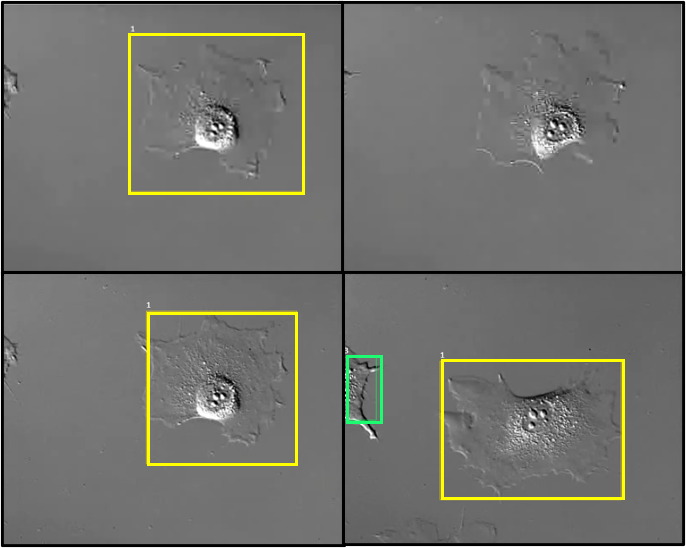}\\
\subcaption{}
\label{study1eg3}
\end{minipage}%
\caption{\textcolor{black}{Poor results collected using the SingSeg microtask using CrowdMOT (top row). (a) Users do not mark \textit{all} objects in the segments. (b) Inconsistent bounding boxes across consecutive segments. (c) Incorrect start frames causing failure with merging annotations across segments when overlapping frames are not annotated. These issues are addressed by using the SingObj microtask design (bottom row)}.}
\label{study1Examples}
\end{figure}

Upon investigation, we identified several factors that caused the annotations to be unsatisfactory with this approach, and illustrate examples in Figure \ref{study1Examples}.  One reason was that workers seldom mark all objects in each segment, as shown in Figure \ref{study1Examples}a. Hence, most objects are not marked in its entirety across the video. Secondly, if an object is marked in two consecutive segments, the bounding boxes drawn by two different workers may vary in size which in turn may not be within the threshold for the merging algorithm to match the two objects. This results in the algorithm mischaracterizing the same object, after merging, for two different objects in the final video. Figure \ref{study1Examples}b illustrates an example of inconsistent boundaries obtained by two workers in two consecutive segments. Finally, errors arise due to incorrect initialization of the objects' start frames. Users often do not mark an object at its first appearance in the video. This problem has higher chances of being compounded with the SingSeg approach, as each segment is given to a different user. In addition, the discontinuity caused by incorrect initialization of the start frame leads to inaccuracy in merging the annotations across segments. Figure \ref{study1Examples}c shows an example in which a worker marked objects at a later frame than their initial appearance.

\hspace{1pt}
\begin{table}[t!]
    \centering
    \begin{tabular}{ l| c |c |c |c } 
    \toprule
    %\multirow{Dataset & Tool & AUC}\\& mean \\
    \multirow{2}{*}{Dataset} & \multirow{2}{*}{Tool} & AUC & TrAcc & Precision\\ & &
    Mean $\pm$ Std & Mean $\pm$Std & Mean$\pm$ Std\\
   % Dataset & Tool & \multicolumn{2}{c}{AUC} \\ mean & std & TM & LA \\
    \toprule
    \multirow{2}{*}{All} %& VATIC & \\%0.06 & 0.42 & 0.03 \\ 
 & CrowdMOT-SingSeg & 0.20 $\pm$ 0.10 &  0.57 $\pm$ 0.18 & \textcolor{black}{0.34 $\pm$ 0.23} \\%0.16 &0.52 & 0.18\\ 
& CrowdMOT-SingObj &  \textbf{0.54}$\pm$0.14 & \textbf{0.96}$\pm$0.05 & \textbf{0.77} $\pm$ 0.26\\
    \midrule
    \multirow{2}{*}{Cell} 
    &  CrowdMOT-SingSeg  & \textcolor{black}{0.16$\pm$0.08} & \textcolor{black}{0.43$\pm$0.08} & \textcolor{black}{0.15$\pm$0.08} \\
    &  CrowdMOT-SingObj &  \textbf{0.55}$\pm$ 0.18 & \textbf{0.95}$\pm$0.05 & \textbf{0.62}$\pm$0.27\\ 
     
    \midrule
    \multirow{2}{*}{Familiar} 
    
    & CrowdMOT-SingSeg & \textcolor{black}{0.23$\pm$0.11} & \textcolor{black}{0.70$\pm$0.12} & \textcolor{black}{0.53$\pm$0.16}\\
    & CrowdMOT-SingObj & \textbf{0.52}$\pm$0.07 & \textbf{0.97}$\pm$0.05& \textbf{0.93}$\pm$0.10 \\ 
    \midrule
    \end{tabular}
    \caption{\textcolor{black}{Performance scores across different types of videos using CrowdMOT-SingSeg and CrowdMOT-SingObj design. CrowdMOT-SingObj outperforms other alternatives by a considerable margin in terms of all three metrics. AUC reflects the accuracy of the size of bounding boxes, TrAcc measures the object's lifetime in the video, and Precision measures the accuracy of the central location of the bounding box. (Std = Standard Deviation)} }  
    \label{study1Tab}
\end{table}

Overall, we observe considerable improvement by using \textbf{CrowdMOT-SingObj}.  Higher scores are observed for all three evaluation metrics.  The higher AUC and Precision scores indicate that CrowdMOT-SingObj is substantially better for tracking both the bounding boxes and their center locations. The higher TrAcc scores demonstrate that it better captures each object's trajectory across its entire lifetime in the video.   

{\color{black}
We found the three flaws that were pointed out for the SingSeg design (illustrated in Figure \ref{study1Examples}) were minimized using the SingObj design. Specifically, (1) we avoid collecting incomplete annotations of objects by simplifying the task and asking one worker to mark one object for the entire video; (2) relying on one worker for the entire video, in turn, leads to a consistent boundary box size of the object across the entire video, and (3) the approach reduces the number of incorrect initializations of the first frame since videos are no longer split into segments.

The most significant lingering issue that we observed is that some workers annotated the same object that was already annotated, despite our explicit instructions to annotate a different object than the one shown. By collecting redundant results, we minimized the impact of this issue. \textcolor{black}{Still, out of a total of 100 parent objects (with their additional 21 children cells), 9 objects (with their additional 7 children cells) were left unannotated because we stopped the propagation of HITs if all three annotators repeated annotation of previously annotated objects.}  A valuable step for future work is to enforce that each annotator selects a distinct object than those previously annotated.

\subsubsection{Results - Human Effort:} 
For \textbf{CrowdMOT-SingSeg}, 25 crowdworkers spent a total of 1,220 minutes (20 hours) to complete all 261 tasks using this system. On average, it took 4.5 minutes and 4.9 minutes to annotate a segment in the cell-based and people-based videos, respectively.  19 unique workers completed 186 tasks on cell videos and 13 workers completed 75 tasks related to familiar videos.  In total, workers annotated 1,567 frames, which averages to 17 key frames out of 320 frames.\footnote{Our analysis is based on a single annotation per segment rather than all three crowdsourced results per segment.} This accounts for about 6.6\% of the total number of video frames (23,938). Of these, 1,050 key frames belong to cell videos and the remaining 517 belong to familiar videos.

For \textbf{CrowdMOT-SingObj}, 42 crowdworkers spent 1,627 minutes (approximately 27 hours) to complete 273 tasks. The average time to annotate an object was 6.9 minutes for the cell videos and 5.1 minutes for the familiar videos.  28 workers completed 126 jobs to annotate the 10 cell videos. 26 workers completed 147 tasks created for the 10 familiar videos. Crowdworkers annotated roughly 8.3\% of the total number of frames (23,938); i.e., 1,994 frames.\footnote{Our analysis is based on a single annotation per object rather than all three crowdsourced results per object.} The remaining frames were interpolated. Of these, 1,151 key frames belonged to cell videos and the remaining 843 were annotated in the familiar videos. Per video (i.e., typically 1,196 frames), a worker annotated, on average, 19 key frames. When comparing the time taken using both strategies, the SingObj strategy appears to take slightly more effort for annotating all the objects in a video. 

In terms of wage comparison, for our collection of videos, both SingObj and SingSeg resulted in a similar number of total jobs (273 versus 261) and total cost (\$136.5 versus \$130.5). However, we observed a considerable difference in the distribution of workload and hence in the annotation performance. In the SingSeg design, while some segments may require a worker to annotate multiple objects, other segments may only consist of one object for annotation. This leads to variability of time and effort a worker would have to spend on the task for different segments of videos. The SingObj framework appears to better scope the time and effort required by a worker by limiting the annotation task to one object for all workers. This is especially important in collaborative crowdwork as workers can often see previous workers' annotations and be able to judge if their effort is equitable. This information may have an impact on their motivation and performance \cite{d2019paying}. Still exceptions exist for both designs.  For the SingSeg design, the last segment can be shorter than previous segments. For SingObj, a worker has to annotate a cell and all its progeny.}

\subsection{Study 2: Effect of Iteration on Video Annotation}

We next examine the influence of iterative tasks on MOT performance for the CrowdMOT-SingObj design. To do so, we evaluate the quality of the annotations when a crowdworker does versus does not observe other object tracking results on the same video. 

\subsubsection{CrowdMOT Implementation:}
We deployed the same crowdsourcing system design as used in study 1 for the CrowdMOT-SingObj microtask design. We assigned each HIT to five workers, and evaluated two rounds of consecutive HITs as described in Steps 1 and 3 below. %We now describe the steps taken to conduct this study below, and focus on the performance of crowdworkers in Steps 1 and 3 to assess the effect of creating iterative tasks:
\begin{quote}
\begin{itemize}
    \item \textbf{Step 1}: We conducted the first round of HITs on all 66 videos, in which workers were asked to annotate only one object per video. The choice of which object to annotate was left to the worker's discretion. We refer to the results obtained on this set of videos as \textit{\textcolor{black}{NonIterative}}.
    \item\textbf{Step 2:} After retrieving the results from step 1, we chose those videos in which workers did a good job in tracking an object for use in creating subsequent tasks. To do so, we emulated human supervision by excluding the videos with an AUC score less than 0.4, which indicates that they have poor tracking results. We refer to the remaining list of videos with good tracking results as \textit{NonIterative-Filtered}. These filtered videos are used in successive tasks for workers to build on the previous annotations.
    \item\textbf{Step 3}: For the second round of HITs, we used all the videos from the NonIterative-Filtered list (i.e., Step 2 results), because they each consist of good tracking results. Workers were shown the previous object tracking results (i.e., that were collected in Step 1) and asked to choose another object for their task that was not previously annotated in the video. We refer to the results obtained in this set of videos as \textcolor{black}{\textit{Iterative}}. 
    \item\textbf{Step 4}: 
    We finally identified those videos from the Iterative HIT (i.e., Step 3 results) that contained good results and so are suitable for further task propagation. To do so, as done for Step 2, we again emulated human supervision by excluding the videos with an AUC score less than 0.4. We refer to the remaining list of videos as \textcolor{black}{\textit{Iterative-Filtered}}.
    \end{itemize}
\end{quote}

\subsubsection{Dataset:} We conducted this study on a larger collection of 116,394 frames that came from 66 live cell videos from the CTMC Cell Motility dataset \cite{Anjum_2020_CVPR_Workshops}. The average number of frames per video is 1,764. The number of cells in the videos vary between 3 to 185.  

\subsubsection{Evaluation Metrics:} We used the same evaluation metrics as used in study 1. Specifically, the quality of crowdsourced results were evaluated using the following three metrics: AUC, \textcolor{black}{TrAcc} and \textcolor{black}{Precision}.  In addition, human effort was calculated in terms of number of key frames annotated per HIT and the time taken to complete each HIT.

\begin{table}[b!]
    \centering
    \begin{tabular}{l c |c |c  }
    \toprule
    %\multirow{} 
     & AUC & TrAcc &Precision\\ & 
    Mean$\pm$Std & Mean$\pm$Std & Mean$\pm$Std\\
    \toprule
         \textcolor{black}{NonIterative} HIT  & 0.50  $\pm$0.14  & \textbf{0.98}$\pm$0.04 &0.47 $\pm$0.29\\
         %\textcolor{black}{NonIterative} HIT- Filtered (43) & 0.57 $\pm$0.09& 0.97 $\pm$0.05 & 0.62 $\pm$0.22\\
         Iterative HIT & \textbf{0.58}$\pm$0.14& 0.97 $\pm$0.05 &\textcolor{black}{\textbf{0.53}$\pm$0.32}\\
         %\textcolor{black}{Iterative} HIT - Filtered (41) & 0.59$\pm$0.12 & 0.98 $\pm$0.05 & 0.54$\pm$0.31\\
         \midrule
    \end{tabular}
    \caption{ \textbf{Analysis of \textcolor{black}{iterative} effect.}  Performance scores for \textcolor{black}{NonIterative} and \textcolor{black}{Iterative} HITs. The filtered list contains videos with AUC $\geq$ 0.4. AUC and Precision scores obtained with Iterative are better than NonIterative showing that iterative tasks has a positive impact on the performance (TrAcc score remains consistent across both cases). }
    \label{avgScores}
\end{table}

\begin{figure}[b!]
\centering
\begin{minipage}[b]{0.55\textwidth}
\includegraphics[width=1\textwidth]{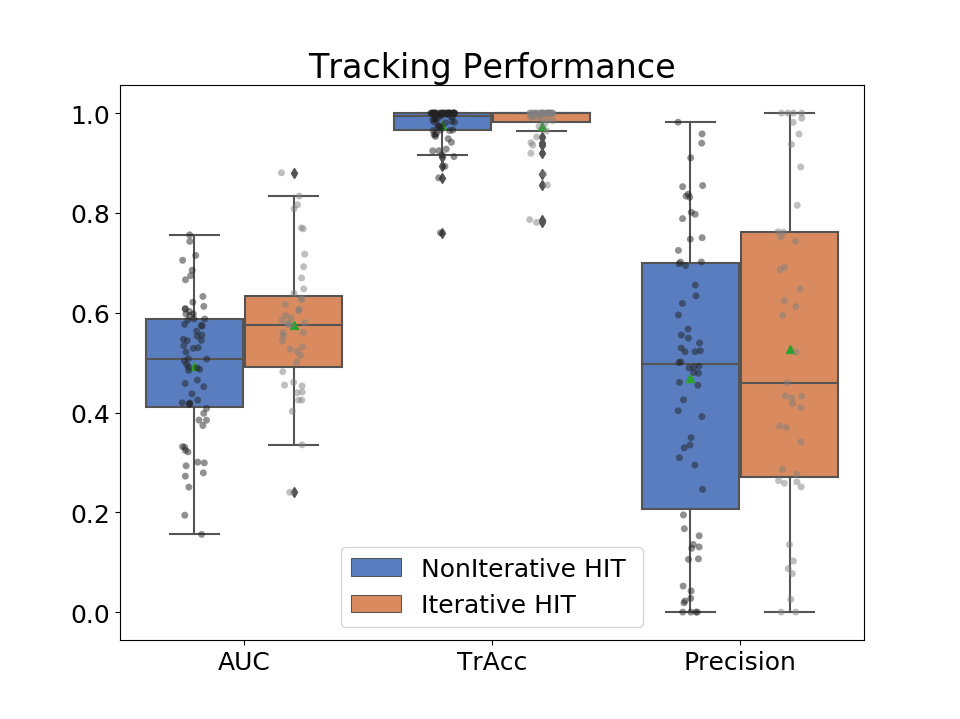}
\subcaption{a.}
\end{minipage}%
\begin{minipage}[b]{0.55\textwidth}
\includegraphics[width=1\textwidth]{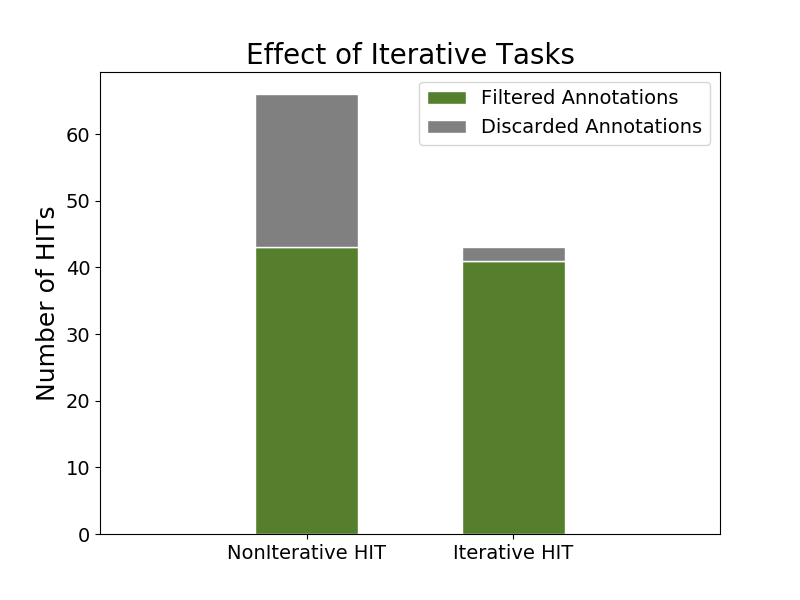}
\subcaption{b.}
\end{minipage}
\caption{\textbf{Analysis of \textcolor{black}{iterative} effect.} (a) Tracking performance of CrowdMOT-SingObj compared across two consecutive rounds of HITs on 66 cell videos. \textcolor{black}{AUC reflects the accuracy of the size of bounding boxes, \textcolor{black}{TrAcc} measures the object's lifetime in the video, and Precision measures the accuracy of the central location of the bounding box.} (b) The filtered list contains videos with AUC $\geq$ 0.4 while the rest are discarded. Fewer number of videos were discarded from the \textcolor{black}{Iterative} HIT results showing that effect of \textcolor{black}{iterative tasks} has a positive impact on the worker performance. }
\label{boxplot}
\end{figure}

\subsubsection{Results - Work Quality:}
We compare the results obtained in NonIterative HITs (Step 1) with those obtained in Iterative HITs (Step 3). Table \ref{avgScores} shows the average AUC, TrAcc, and Precision scores, while Figure \ref{boxplot} shows the distribution of these scores. For completeness, we include the scores across all four sets of videos described in Steps 1-4 in Appendix (Table \ref{study2results_extended} and Figure \ref{study2fig_extended}). 

As shown in Table \ref{avgScores}, we found that workers performed better in the Iterative HITs as compared to the NonIterative HITs. Observing this overall improvement of the worker performance, we hypothesize that the existing annotation may have helped guide the workers of the second HIT to better understand the requirements and annotate accordingly. This improvement occurred despite the fact that sample videos already were provided in the instructions to show how to annotate using the tool for both scenarios (\textcolor{black}{Iterative} and \textcolor{black}{NonIterative}), as mentioned in Section \ref{Approach}. This suggests that observing a prior annotation on the same video offers greater guidance than only having access to a video within the instructions.
 
 Our findings lead us to believe that \textcolor{black}{observing previous annotations} has more of an impact on the resulting size of the bounding box than the center location of the box.  After excluding annotations with low AUC scores (i.e., AUC $<$ 0.4), we found that from the NonIterative HIT, 43 out of 66 videos consisted of satisfactory annotations, which accounted for 65\%. However, using the same cutoff score of AUC with the Iterative HIT annotations, crowdworkers provided significantly better results on 41 out of 43 videos (p value = 0.0020 using Student's t-Test). This resulted in 95\% of the videos achieving better annotations. There was a slight improvement in the \textcolor{black}{Precision} scores as well, though it was not significant. 
 
 Across both NonIterative and Iterative results, the average \textcolor{black}{TrAcc} scores were above 97\%. This shows that, for both approaches, workers persisted and remained reliable in terms of marking the object through its lifetime in the video. While the \textcolor{black}{TrAcc} scores of annotations were generally high, the cause for some objects scoring low was attributed to the temporal offset of frames for objects that underwent either a splitting or left the frame. 
 
 Next, we referred to the difficulty level of the videos to assess the impact of the varying difficulty levels. Specifically, we leveraged the difficulty ratings of the videos provided by our in-house annotators, which was assigned during ground truth generation. We observed that out of 28 easy, 24 medium and 14 hard videos, those that were removed in the first round included 5 easy, 10 medium and 8 hard videos. While, predictably, more easy videos passed to the second round of HITs, we also note that the second round consisted of about 50\% of the videos belonging to medium/hard categories. In other words, workers in the second round were asked to annotate videos from a mixed bag of all three difficulty levels.

\subsubsection{Results - Human Effort:}
A total of 78 unique workers completed the 545 tasks for the 66 videos. 24 unique workers participated in the \textcolor{black}{NonIterative} HITs and 66 unique workers participated in the \textcolor{black}{Iterative} HITs, with 12 workers participating on both sets of HITs.

Crowdworkers annotated a total of 2,468 frames, which is about 2.1\% of the total number of frames, with an average of 14 key frames per object in a video \footnote{Our analysis is based on a single annotation per object rather than all five crowdsourced results per object.}. Of these, 647 key frames were annotated for the \textcolor{black}{NonIterative} HITs, while 1,761 key frames were annotated for the \textcolor{black}{Iterative} HITs. This suggests that workers that were shown prior annotations invested more effort into submitting high quality annotations.  This finding is reinforced when examining the time taken to complete the tasks.  On average, \textcolor{black}{NonIterative} tasks took 4.6 minutes per object, while the \textcolor{black}{Iterative} tasks were completed in 5.6 minutes. \footnote{Overall, we found the time taken by crowdworkers to annotate each object using CrowdMOT-SingObj is consistent with that in study 1, with the average being 5.02 minutes per job.}  
\section{Discussion}

\subsection{Implications}
We focused on designing effective strategies for employing crowdsourcing to track multiple objects in videos.  Rather than relying upon expert workers~\cite{vondrick_efficiently_2013}, which ignores the potential of a large pool of workers, our strategies aim at leveraging the non-expert crowdsourcing market by (1) designing the microtask to be simple (i.e. SingObj) and (2) providing additional guidance in the form of prior annotations (i.e. iterative tasks).  Our experiments reveal benefits of implementing these two strategies when collecting annotations on a complex task like tracking multiple objects in videos, as discussed below.  

Our first strategy \textbf{decomposes the video annotation task into a simpler microtask} in order to facilitate gathering better results. Our experiments show that simplifying the annotation task by assigning a crowdworker to the entire lifetime of one object in a video yielded a higher quality of annotations than assigning a crowdworker ownership for all objects over the entire segment of the video. Subsequent to our analysis, we learned that our findings complement those found in studies that examine the human attention level in performing the MOT task. For instance, prior work showed that human performance decreases for object tracking as the number of objects grows and the best performance was shown in the annotation of first object \cite{alvarez2007many, holcombe2012exhausting}.  Another study showed that humans can track up to four objects in a video accurately \cite{intriligator2001spatial}. An additional study demonstrated that tracking performance is dependent on factors such as object speed as well as spatial proximity between objects \cite{alvarez2007many}. For example, they showed that if the objects were moving at a sufficiently slow speed, humans could track up to eight objects. While our experiments demonstrate that SingObj ensures higher accuracy, further exploration could determine the limits of the conditions under which SingObj is preferable (e.g., possibly for a large number of objects). 

Our second strategy demonstrates that \textbf{exposing crowdworkers to prior annotations by creating iterative tasks can have a positive influence} on their performance. Having an interactive workflow of microtasks through consecutive rounds of crowdsourcing, can provide crowdworkers a more holistic understanding of how their work contributes to the bigger goal of the project. This, in turn, can improve the quality of their performance, as noted by related prior work  \cite{bigham2015human}. 

More broadly, this work can have implications in designing crowdsourcing microtasks for other applications that similarly leverage spatial-temporal information, such as ecology \cite{estes2018spatial}, wireless communications \cite{raleigh1998spatio}, and tracking tectonic activity \cite{miller2006spatial}. Similar to MOT, these applications can also choose to decompose tasks either temporally (SingSeg) or spatially (SingObj).  Our findings paired with the constraints imposed from our experimental design, with respect to the length of videos and total number of objects, underscore certain conditions for which we anticipate the SingObj design will yield higher-quality and more consistent results. 

By \textbf{releasing the CrowdMOT code publicly, we aim to encourage MOT crowdsourcing in a greater diversity of domains, including data that is both familiar and typically unfamiliar to lay people}. Much crowdsourcing work examines involving non-experts to annotate data that is uncommon or unfamiliar to lay people. For example, researchers have relied on crowdsourcing to annotate lung nodule images \cite{boorboor_crowdsourcing_2018}, colonoscopy videos \cite{park2017crowdsourcing}, hip joints in MRI images \cite{chavez2013crowdsourcing}, and cell images~\cite{gurari2014use,gurari2015collect,sameki2015characterizing,gurari2016investigating}.  The scope of such efforts has been accelerated in part because of the Zooniverse platform, which simplifies creating crowdsourcing systems \cite{borne2011zooniverse}. Our work intends to complement this existing effort and anticipates users may benefit from using CrowdMOT to crowdsource high-quality annotations for their biological videos showing cells (that exhibit a splitting behavior).  Providing a web version of the tool empowers users and researchers to more easily  annotate videos by reducing the overhead of tool installation and setup. This can be valuable for many potential users, especially those lacking domain expertise. 

\subsection{Limitations and Future Work}

\textcolor{black}{An important issue we observed with the annotations was users are often unable to annotate the object from the correct starting frame. A useful enhancement would be to ease this process by having an algorithm seed each object in the first frame it appears, and thereby guide the worker into annotating that object only. }

While we are encouraged by the considerable improvements in the tracking annotation results obtained from workers using our system, it is possible \textbf{more sophisticated interpolation schemes} could lead to further improvements. The current framework uses linear interpolation to fill the intermediate frames between user-defined key frames with annotations, as it was similarly used in popular video annotation systems \cite{vondrick_efficiently_2013, yuen2009labelme}.  An interesting area for future work is to explore how changing the interpolation schemes (e.g., level set methods) will impact crowdworker effort and annotation quality. In the future, we also plan to increase the size of our video collection to assess the versatility of our framework on different types of videos. 

Although our crowdsourcing approach yields significant improvements, \textbf{future work is needed to address certain settings where we believe this approach may not be viable}. One example is for very long videos, since every user has to watch the entire video to mark one object. In such scenarios, it may be beneficial to design a microtask that can integrate both SingObj and SingSeg strategies. For example, a long video can be divided into smaller segments, where each worker can then be asked to annotate one object per segment. In addition, our current framework supports objects that split into two children, like in the case of cells in biomedical research. This can be extended to support objects that undergo any number of splits, for example, in videos depicting ballistic testing that involve an object breaking into multiple pieces. Finally, future work will need to examine how to generalize MOT solutions for videos that show 10s, 100s, or more of objects that need to be tracked.
 
\section{Conclusion}
We introduce a general-purpose crowdsourcing system for multiple object tracking that also supports lineage tracking.  Our experiments demonstrate significant flaws in the existing state-of-the-art crowdsourcing task design.  We quantitatively demonstrate the advantage of two key micro-task design options in collecting much higher quality video annotations: (1) have a single worker annotate a single object for the entire video and (2) show workers the results of previously annotated objects on the video.  To encourage further development and extension of our framework, we will publicly share our code.

\section*{Acknowledgements}
This project was supported in part by grant number 2018-182764 from the Chan Zuckerberg Initiative DAF, an advised fund of Silicon Valley Community Foundation.  We thank Matthias M{\"{u}}ller, Philip Jones and the crowdworkers for their valuable contributions in this research.  We also thank the anonymous reviewers for their valuable feedback and suggestions to improve this work.

\bibliographystyle{ACM-Reference-Format}
\bibliography{references}

\appendix
\section*{Appendix}
 \label{section: appendix}
This section includes supplementary material to Sections \ref{sec:vatic_evaluation} and \ref{section: experiments}.

\begin{itemize}
    \item Section \ref{sec: pilot_crowdmot} provides details of a pilot study conducted using the CrowdMOT-SingObj framework, which was used for comparison to the VATIC system analyzed in Section \ref{sec:vatic_evaluation}.
    \item We report evaluation results that supplement the analysis conducted in study 2 of Section \ref{section: experiments}.
\end{itemize}
\section{Pilot Study: Evaluation of CrowdMOT}
\label{sec: pilot_crowdmot}

This pilot study was motivated by our observation that we received poor quality results from using the state-of-art crowdsourcing system, VATIC (Section \ref{sec:vatic_evaluation}), which employs the SingSeg design.  We conducted a follow-up pilot experiment to assess the quality of results obtained using our alternative design: CrowdMOT with the SingObj design.  We observed a considerable improvement in the quality of results using our CrowdMOT-SingObj system over those obtained with VATIC (Section \ref{sec:vatic_evaluation}), which led us to conduct subsequent experiments (Section \ref{section: experiments}).

\subsection{Experimental Design} 

\textit{Dataset.} We conducted this study on the same dataset as used in Section \ref{sec:vatic_evaluation} which included 35 videos containing 49,501 frames, with 15 videos showing familiar content (i.e. people) and the remaining 20 videos showing unfamiliar content (i.e. cells).\\

\noindent\textit{CrowdMOT Configuration.} We deployed the same crowdsourcing design as used in Study 1 of Section \ref{section: experiments} for the CrowdMOT-SingObj microtask design. We created a new HIT for each object.  Workers were asked to mark only one object in the entire video and could only submit the task after both detecting and tracking an object. Our goal was to study the trend of the worker performance, so we collected annotations for only two objects per video rather for all objects in the entire video.  
 
 Each HIT was assigned to five workers. This resulted in a total of 350 jobs for 35 videos. Out of the five annotations, we picked the annotation with the highest AUC score per video to use as input for the subsequent, second posted HIT. This choice of using the annotations with the highest AUC score as input is intended to simulate a human supervision of selecting the best annotation. Evaluation was conducted for the two objects and any of their progeny for all videos.\\

\noindent\textit{Crowdsourcing Environment and Parameters.} As was done in Section \ref{sec:vatic_evaluation}, we employed crowdworkers from Amazon Mechanical Turk (Turk) who completed at least 500 HITs and had at least 95$\%$ approval rating. Each worker was paid $\$$0.50 per HIT and given 30 minutes to complete that HIT.\\  

\noindent\textit{Evaluation Metrics.} We used the same three metrics as used in Section \ref{sec:vatic_evaluation} to evaluate the results, namely, AUC, TrAcc, and Precision.

\subsection{Results}

We observe considerable improvement using CrowdMOT with the SingObj microtask design compared to the VATIC; i.e., it results in higher scores for all three evaluation metrics. Specifically, with CrowdMOT-SingObj, the AUC score was 0.50, TrAcc was 0.96, and Precision was 0.63, as compared to 0.06, 0.42 and 0.03 with VATIC. The higher AUC and Precision scores indicate that CrowdMOT-SingObj is substantially better for tracking both the bounding boxes and their center locations. The higher TrAcc scores demonstrate that it better captures each object’s trajectory across its entire lifetime in the video. 

\section{Evaluation of NonIterative versus Iterative Task Designs}

\begin{figure}[h!]
    \centering
    \includegraphics[width=0.7\textwidth]{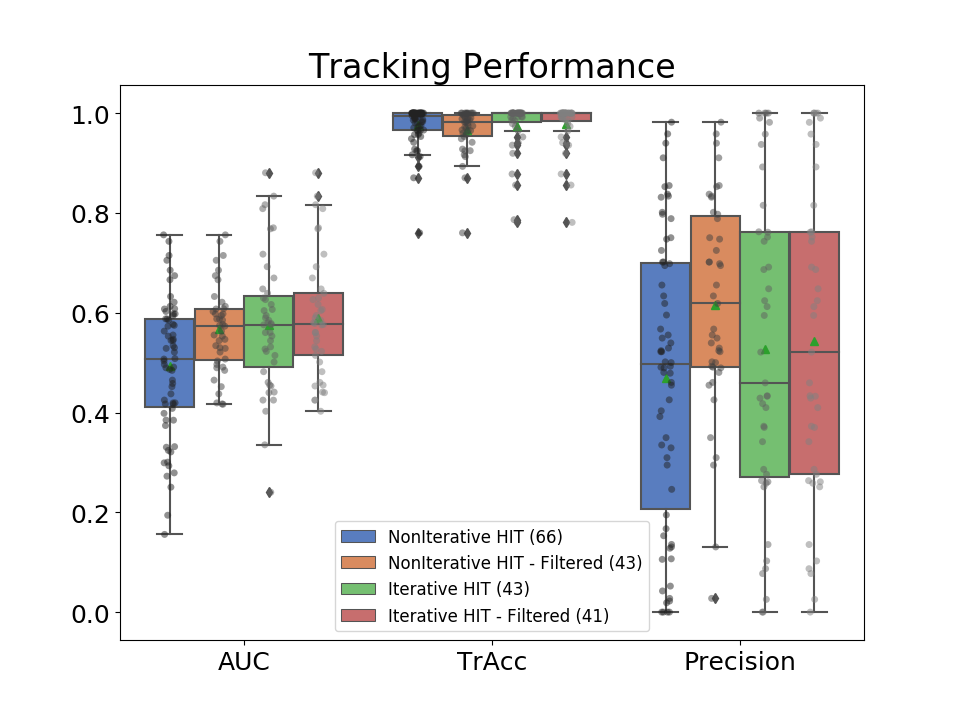}
    \caption{\textbf{Analysis of iterative effect (Study 2)}. Tracking performance of CrowdMOT compared across the four groups of videos described in Study 2, with the values in the parentheses representing the number of videos in each case. AUC reflects the accuracy of the size of bounding boxes, TrAcc measures the object’s lifetime in the video, and Precision measures the accuracy of the central location of the bounding box. Iterative HITs obtain higher quality results than NonIterative HITs.}
    \label{study2fig_extended}
\end{figure}

Supplementing Section 6.2 of the main paper, we report additional results comparing the performance of our NonIterative versus Iterative task designs. The distribution of AUC, TrAcc and Precision scores for all four sets of videos described in Step 1-4 (NonIterative, NonIterative-Filtered, Iterative and Iterative-Filtered) are illustrated in Figure \ref{study2fig_extended} and the average of those scores are summarized in Table \ref{study2results_extended}. The filtered sets contain those annotations that were of high quality (i.e. AUC $\geq$ 0.4) from the results for the NonIterative and Iterative results respectively. We observe a considerable difference in the scores between the NonIterative results and its corresponding filtered set.  This observation contrasts what is observed for the the Iterative results and its corresponding filtered set. We attribute this distinction to the fact that more annotations are discarded for the NonIterative tasks (i.e 23 out of 66 HITs) than the Iterative tasks (i.e. 2 out of 43 HITs).  This finding offers promising evidence that Iterative tasks yield better results for collecting MOT annotations. 

\begin{table}[t!]
    \centering
    \begin{tabular}{l c |c |c  }
    \toprule
    %\multirow{} 
    & AUC & TrAcc &Precision\\ & 
    \textcolor{black}{Mean$\pm$Std} & \textcolor{black}{Mean$\pm$Std} & \textcolor{black}{Mean$\pm$Std}\\
    %& AUC & TM & LA\\
    \toprule
         \textcolor{black}{NonIterative} HIT (66) & 0.50  $\pm$0.14  & {0.98}$\pm$0.04 &0.47 $\pm$0.29\\
         \textcolor{black}{NonIterative} HIT- Filtered (43) & 0.57 $\pm$0.09& 0.97 $\pm$0.05 & 0.62 $\pm$0.22\\
         \textcolor{black}{Iterative} HIT (43)& \textcolor{black}{{0.58}$\pm$0.14}& 0.97 $\pm$0.05 &\textcolor{black}{{0.53}$\pm$0.32}\\
         \textcolor{black}{Iterative} HIT - Filtered (41) & 0.59$\pm$0.12 & 0.98 $\pm$0.05 & 0.54$\pm$0.31\\
         \midrule
    \end{tabular}
    \caption{ \textbf{Analysis of \textcolor{black}{iterative} effect (Study 2).}  Performance scores for \textcolor{black}{NonIterative} and \textcolor{black}{Iterative} HITs, with the value in the parentheses denoting the total number of videos in each set. The filtered list contains videos with AUC $\geq$ 0.4. AUC and Precision scores obtained with Iterative (third row) are better than NonIterative (first row) showing that iterative tasks has a positive impact on the performance (TrAcc score remains consistent across both cases). Only two videos are discarded from Iterative HIT as compared to the 23 videos filtered out from NonIterative HIT.  }
    \label{study2results_extended}
\end{table}

\end{document}